\documentclass[10pt,twocolumn,letterpaper]{article}

\usepackage{cvpr}              % To produce the CAMERA-READY version
\usepackage{makecell}
\def\paperID{6316} % *** Enter the Paper ID here
\def\confName{CVPR}
\def\confYear{2026}
\usepackage{multirow}
\usepackage[sort&compress]{natbib}
\usepackage{hyperref}
\usepackage{balance}

\title{PGR-Net: Prior-Guided ROI Reasoning Network \\for Brain Tumor MRI Segmentation}

\author{Jiacheng Lu$^{1}$, Hui Ding$^{1*}$, Shiyu Zhang$^{1}$, Guoping Huo$^{2*}$\\
$^{1}$College of Information Engineering, Capital Normal University, Beijing, China\\
$^{2}$School of Artificial Intelligence, China University of Mining and Technology-Beijing, Beijing, China\\
{\tt\small jchengl@foxmail.com, dhui@cnu.edu.cn$^{*}$, sh1yuzh@163.com, kuoping@cumtb.edu.cn$^{*}$}\\
\tt\small $^{*}$Corresponding authors}
\begin{document}
\maketitle
\begin{abstract}
Brain tumor MRI segmentation is essential for clinical diagnosis and treatment planning, enabling accurate lesion detection and radiotherapy target delineation. However, tumor lesions occupy only a small fraction of the volumetric space, resulting in severe spatial sparsity, while existing segmentation networks often overlook clinically observed spatial priors of tumor occurrence, leading to redundant feature computation over extensive background regions.
%%Brain tumor MRI segmentation is essential for clinical diagnosis and treatment planning, enabling accurate lesion detection and radiotherapy target delineation. However, tumor lesions occupy only a small fraction of the volumetric space, resulting in severe spatial sparsity, while existing segmentation networks often overlook clinically observed spatial priors of tumor occurrence and thus waste computation over large background regions. 
To address this issue, we propose PGR-Net (Prior-Guided ROI Reasoning Network)—an explicit ROI-aware framework that incorporates a data-driven spatial prior set to capture the distribution and scale characteristics of tumor lesions, providing global guidance for more stable segmentation.
Leveraging these priors, PGR-Net introduces a hierarchical Top-$K$ ROI decision mechanism that progressively selects the most confident lesion candidate regions across encoder layers to improve localization precision. We further develop the WinGS-ROI (Windowed Gaussian–Spatial Decay ROI) module, which uses multi-window Gaussian templates with a spatial decay function to produce center-enhanced guidance maps, thus directing feature learning throughout the network. With these ROI features, a windowed RetNet backbone is adopted to enhance localization reliability.
Experiments on BraTS-2019/2023 and MSD Task01 show that PGR-Net consistently outperforms existing approaches while using only 8.64M Params, achieving Dice scores of 89.02\%, 91.82\%, and 89.67\% on the Whole Tumor region. Code is available at \url{https://github.com/CNU-MedAI-Lab/PGR-Net}.

\end{abstract}

\section{Introduction}

\begin{figure}[htbp]
    \centering
    \includegraphics[width=0.9\linewidth]{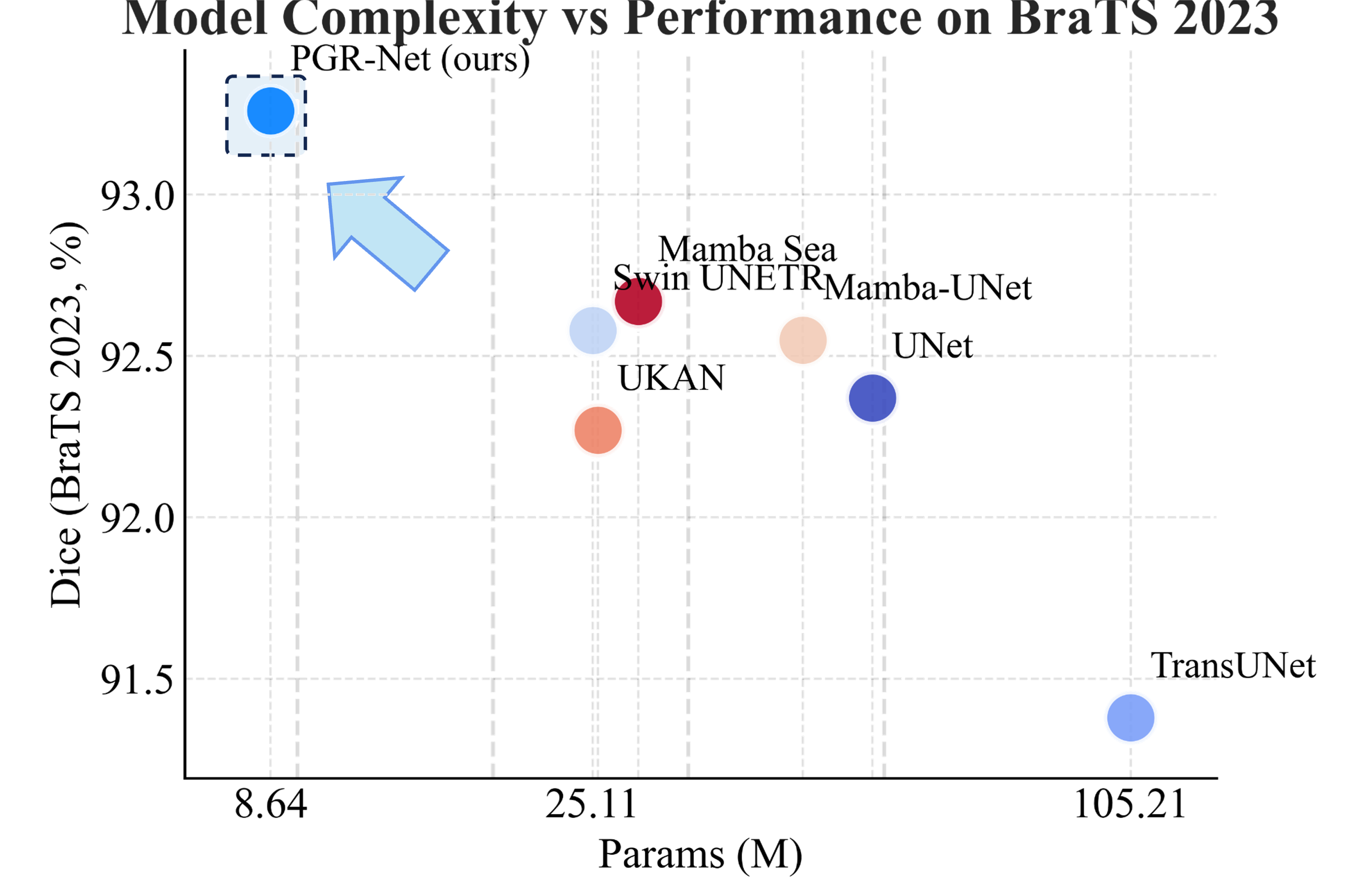}
    \caption{Model complexity vs. BraTS 2023 performance (average Dice over the three tumor regions). }
    \label{fig:abstract}
\end{figure}

\begin{figure}[htbp]
    \centering
    \includegraphics[width=0.76\linewidth]{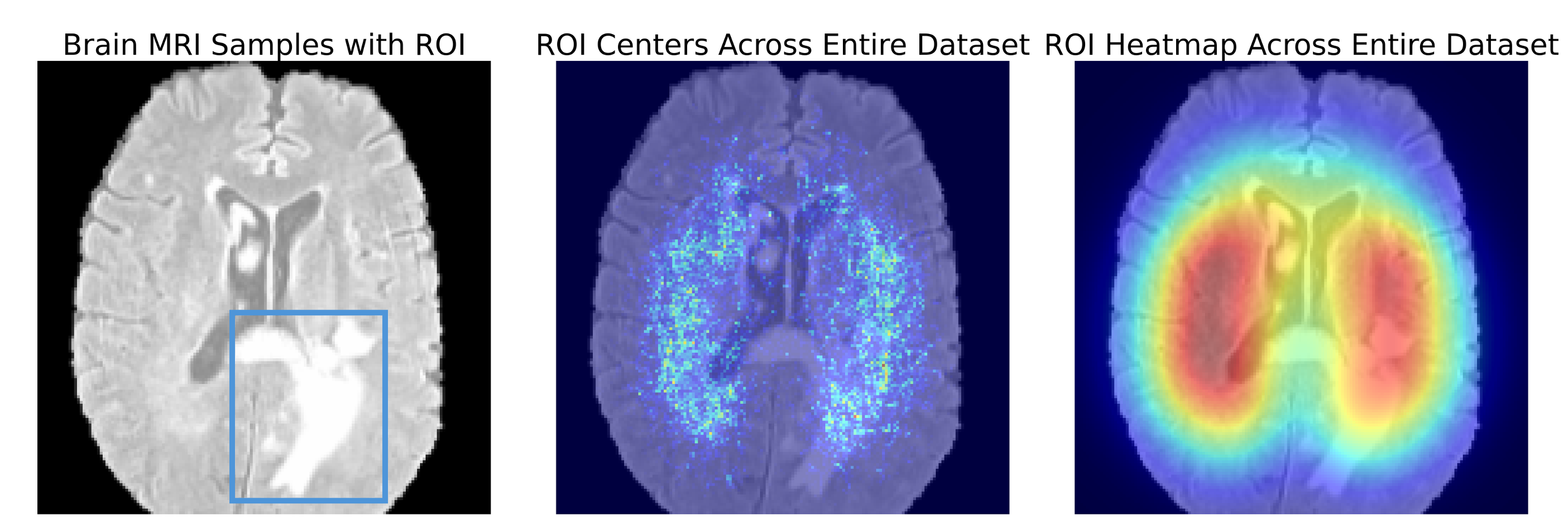}
    \caption{An illustrative example of lesion distribution in the BraTS 2023 dataset. From left to right: a sample MRI image, lesion center distribution map across the entire dataset, and lesion ROI distribution heatmap across the entire dataset.}
    \label{fig:intro}
\end{figure}

% Brain tumor MRI segmentation is essential for clinical diagnosis and preoperative planning, enabling accurate lesion detection, volumetric measurement, and radiotherapy target delineation \cite{Tan1, Liang1}. Deep learning-based methods have achieved remarkable progress, with U-Net \cite{Ron1} serving as the cornerstone of medical image segmentation. Beyond convolutional networks \cite{Fuku}, sequence modeling paradigms such as Transformers \cite{transformer} and Mamba state-space models (SSM) \cite{mamba} further advance segmentation by enhancing attention to organs and lesions. Representative methods include TransUNet \cite{transunet}, UNETR \cite{unetr}, Swin UNETR \cite{swin}, and VMamba-based approaches \cite{vmamba, visionmamba} such as Mamba-UNet \cite{mambaunet} and M-Net \cite{mnet}.
Brain tumor MRI segmentation is essential for diagnosis and preoperative planning, enabling accurate lesion detection, volumetric measurement, and radiotherapy target delineation \cite{Tan1, Liang1}. Deep learning methods have achieved remarkable progress, with U-Net \cite{Ron1} as the cornerstone of medical image segmentation. Beyond convolutional networks \cite{Fuku}, sequence modeling such as Transformer \cite{transformer} and Mamba state-space model (SSM) \cite{mamba} further advance segmentation by enhancing attention to organs and lesions. Representative methods include TransUNet \cite{transunet}, UNETR \cite{unetr}, Swin UNETR \cite{swin}, and VMamba-based approaches \cite{vmamba, liu2024vmamba} such as Mamba-UNet \cite{mambaunet} and M-Net \cite{mnet}.

Although segmentation accuracy has steadily improved, most models still struggle with the spatial sparsity and scale imbalance of brain tumors. In BraTS2023, the average tumor region occupies only about 2,740 pixels—around 10.7\% of the entire image (160×160). This imbalance causes the model to be dominated by background features during early training. In later stages, although the model may roughly localize the tumor, it still consumes substantial computational resources modeling background and healthy tissues, resulting in suboptimal localization accuracy and structural consistency.

Tumor distribution in brain MRI follows statistically regular spatial patterns. As shown in Fig.~\ref{fig:intro} (BraTS2023 dataset), lesion centers predominantly occur near the frontal–temporal junction \cite{cli1, cli2}, while lesions are rarely observed in the occipital lobe \cite{cli3}. However, most deep segmentation models still assume a uniform lesion distribution, ignoring these clinically observed spatial priors, which leads to increased computational redundancy.

To address the above issues, we propose an explicit ROI-aware brain tumor MRI segmentation network, termed PGR-Net (Prior-Guided ROI Reasoning Network). The proposed framework achieves a unified process from statistical modeling to spatial propagation. The main contributions of this work are summarized as follows:

1. We design PGR-Net, an explicit ROI-aware brain tumor segmentation network built upon a windowed RetNet backbone. The network incorporates a hierarchical Top-$K$ ROI selection mechanism that progressively filters high-confidence regions across multiple spatial locations and scales. Differentiated ROI guidance strategies are applied at encoder, decoder, and skip-connections to ensure spatial attention consistency throughout feature propagation.

2. A novel Windowed Gaussian–Spatial Decay ROI (WinGS-ROI) module is proposed, generating center-enhanced spatial guidance maps using multi-window Gaussian templates and spatial decay functions along ROI boundaries, effectively guiding feature learning across all layers of PGR-Net.

3. To support region-aware learning, we construct a generalized ROI prior template set by analyzing the center distribution and scale characteristics of tumors in the training set. Representative distributional peaks are extracted to provide spatial and scale constraints, enabling the model to leverage prior knowledge from early stages of training.

%%1.	By analyzing the center distribution and scale characteristics of all tumors in the training set, we construct a generalized ROI prior template set. This set extracts representative distributional peaks to provide explicit spatial and scale constraints, enabling the model to possess region-aware capabilities from the early training stages.

%%2.	Based on the ROI priors, we design PGR-Net with a windowed RetNet (a Transformer variant) as its visual backbone, incorporating a hierarchical Top-k ROI selection mechanism that progressively filters high-confidence regions across multiple spatial locations and scales. Differentiated ROI guidance strategies are introduced at each stage of the encoder, decoder, and skip connections to ensure spatial attention consistency throughout the feature propagation process.

%%3.	To further enhance the expressiveness of ROI features, we propose the Windowed Gaussian–Spatial Decay ROI (WinGS-ROI) module. This module generates center-enhanced spatial guidance maps using multi-window Gaussian templates and spatial decay functions along ROI boundaries , thereby achieving effective visual feature guidance across all layers of PGR-Net.

Comprehensive experiments on multiple benchmarks verify that the proposed prior-guided strategy and PGR-Net attain state-of-the-art segmentation accuracy with lower computational cost (see Fig.~\ref{fig:abstract}).

\section{Related Work}
\subsection{RetNet}
Transformer provides strong global modeling ability but suffers from high computational cost. To address this, Sun et al. proposed RetNet\cite{retnet}, which replaces multi-head self-attention with a retention mechanism.

Given an input sequence \(\{x_t\}_{t=1}^n\),
\begin{equation}
q_t = W_Q x_t, \quad k_t = W_K x_t, \quad v_t = W_V x_t.
\end{equation}
In recurrent mode, the state evolves as
\begin{equation}
s_t = \gamma s_{t-1} + K_t^\top v_t, \quad
o_t = Q_t s_t,
\end{equation}
where \(\gamma \in (0,1]\) is a decay factor.
In parallel (training) mode,
\begin{equation}
\mathrm{Retention}(X) = (Q K^\top \odot D) V,
\end{equation}
where D encodes causal masking and exponential decay.
RetNet and its visual variant\cite{rmt} improve efficiency and stability for high-resolution vision tasks.
\subsection{Region-aware Segmentation}

In medical image segmentation, lesion localization remains a key challenge. Early approaches\cite{cas1, cas2, cas3} adopted two-stage pipelines that first detect lesions and then perform fine-grained segmentation, but suffer from stage separation and error accumulation. Recent methods\cite{attention1, attention2, attention3} use convolutional or Transformer-based attention to emphasize lesion regions. However, they largely ignore lesion spatial distribution patterns and implicitly assume uniform distributions, leading to redundant feature computation.

Some works\cite{roi1, roi2} introduce prior ROI guidance to improve spatial awareness, but such hard-guided strategies generalize poorly because they fail to capture the underlying ROI distribution patterns, limiting their effectiveness for region-aware segmentation.

\section{Method}
\subsection{ROI Prior Construction}
\begin{figure}[htbp]
    \centering
    \includegraphics[width=0.7\linewidth]{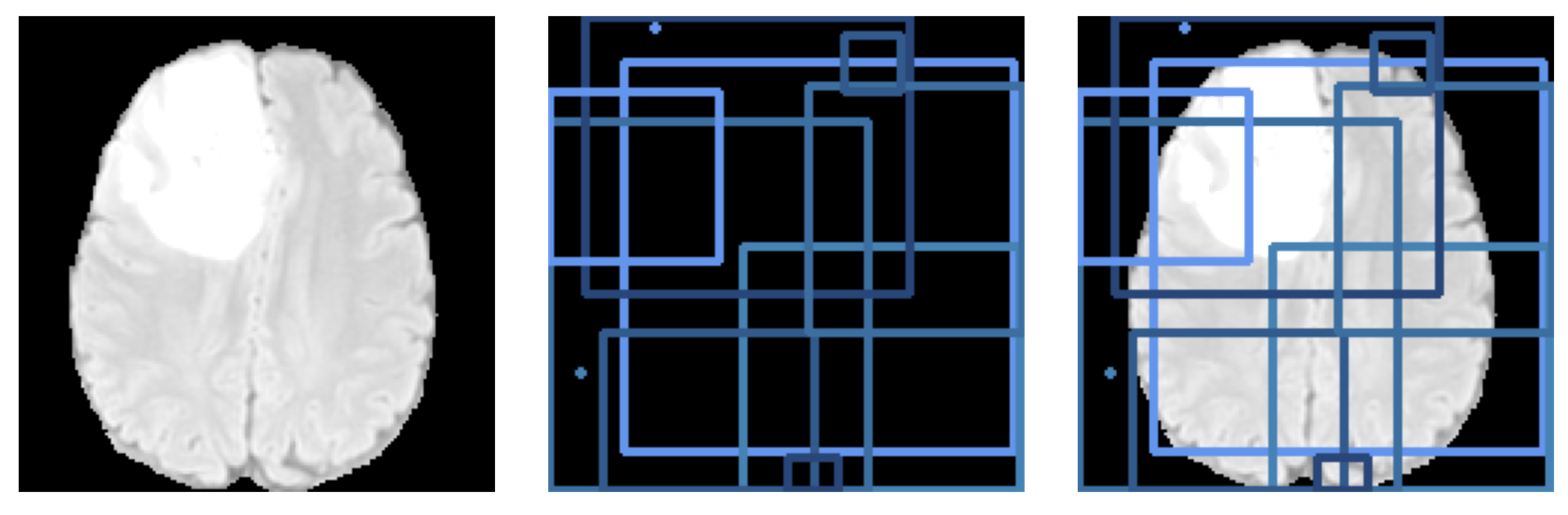}
    \caption{Example templates from BraTS 2023 training set: sample FLAIR image, extracted ROI, and ROI overlaid on the image.}
    \label{fig:rois}
\end{figure}

We construct ROI prior templates (Fig.~\ref{fig:rois}) from the training set to provide consistent spatial and scale guidance:
\begin{equation}
\mathcal{R}_0 = \{ (r_i, c_i) \}_{i=1}^N,
\end{equation}
where $r_i$ is the normalized ROI scale ratio and $c_i$ is the representative center.

Given $M$ mask samples $\{Y_m\}_{m=1}^M$, connected components are extracted:
\begin{equation}
\mathcal{B}_m = \{ B_{m,j} \mid j = 1, \ldots, N_m \},
\end{equation}
where $B_{m,j}$ denotes the $j$-th connected component in sample $Y_m$.

For each component, compute the minimum bounding rectangle with height $h_{m,j}$ and width $w_{m,j}$, side length:
\begin{equation}
s_{m,j} = \max(h_{m,j}, w_{m,j}),
\end{equation}
and center coordinates:
\begin{equation}
(x_{m,j}, y_{m,j}) = \left( \frac{x_{\min} + x_{\max}}{2}, \frac{y_{\min} + y_{\max}}{2} \right).
\end{equation}

\begin{figure}[htbp]
    \centering
    \includegraphics[width=0.9\linewidth]{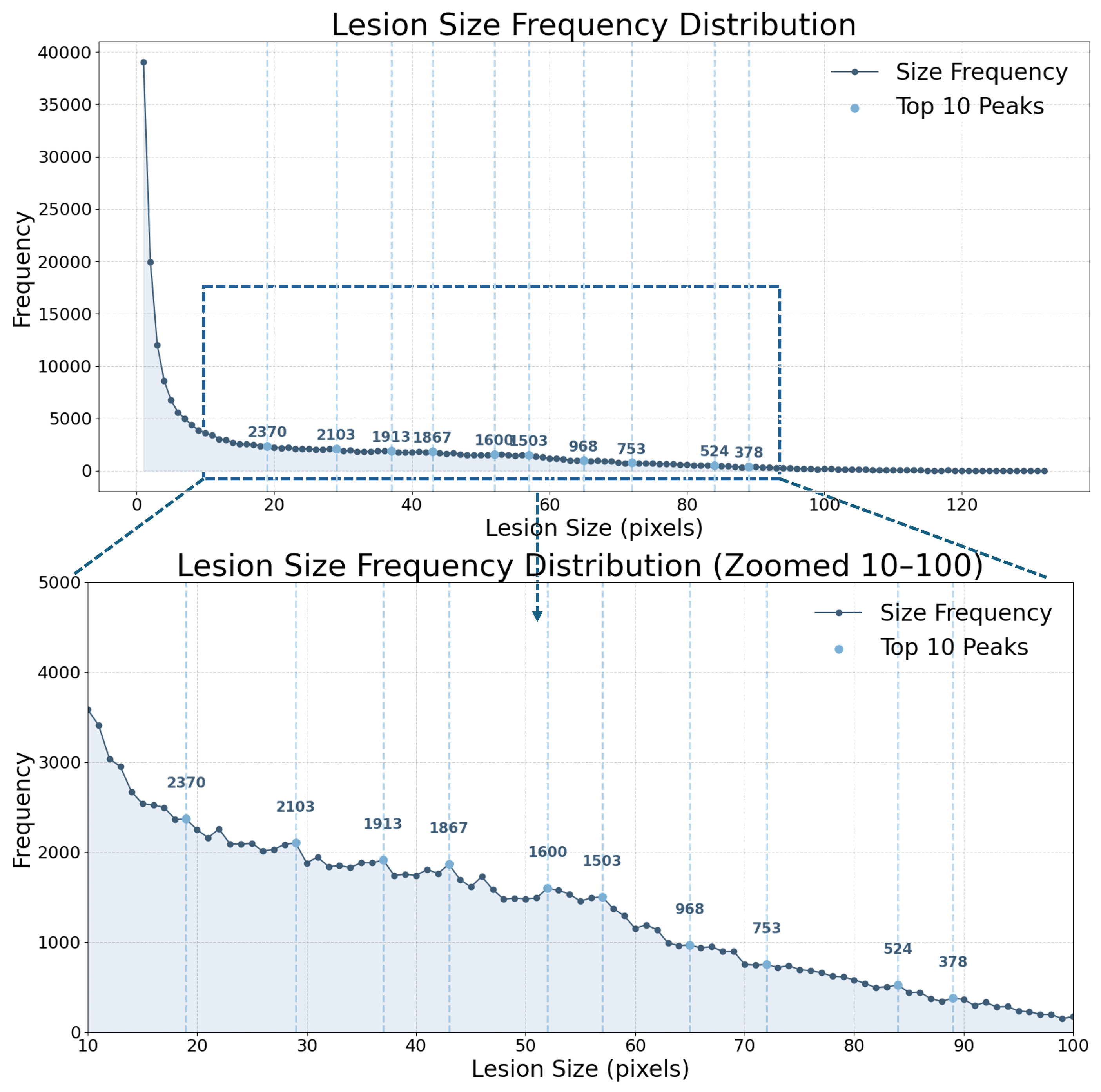}
    \caption{Lesion size distribution of the BraTS 2023 training set, with lesion size on the horizontal axis and frequency on the vertical axis. The lower plot shows a magnified view of the 10–90 range with annotated 10 peaks.}
    \label{fig:roi_curve}
\end{figure}

Components with $s_{m,j} < s_\text{min}$ are discarded (e.g., 10). All valid sizes are collected:
\begin{equation}
\mathcal{S} = \{ s_{m,j} \mid s_{m,j} \ge s_\text{valid} \},
\end{equation}
where $s_\text{valid}$ is a lower bound for valid sizes(e.g., 20). The scale distribution is below, $\delta(\cdot)$ is the delta Dirac function.
\begin{equation}
P(s) = \frac{1}{|\mathcal{S}|} \sum_{s' \in \mathcal{S}} \delta(s - s').
\end{equation}
Local maxima are detected from the scale distribution with a minimum spacing of $d_\text{min}$ (e.g., $d_\text{min}=5$), and the top $N$ peaks (e.g., Fig.~\ref{fig:roi_curve}) are normalized to the image height:
\begin{equation}
r_i = \frac{s_i}{H}, \quad i = 1, \ldots, N.
\end{equation}

These peaks provide candidate lesion scales. For each scale $r_i$, centers of components within the corresponding peak cluster are collected and averaged to obtain a representative center $c_i$:
\begin{equation}
c_i = \frac{1}{|\mathcal{C}_i|} 
\sum_{(x_{m,j}, y_{m,j}) \in \mathcal{C}_i} 
\left( \frac{x_{m,j}}{W}, \frac{y_{m,j}}{H} \right).
\end{equation}

Spatial clustering is further constrained by a neighborhood radius $d$ (set to $30$), so only ROIs with nearby centers are grouped together, preventing distant components with similar sizes from being merged.

% \section{Method}
\subsection{Proposed Prior-Guided ROI Reasoning Network}

\begin{figure*}[htbp]
    \centering
    \includegraphics[width=0.9\linewidth]{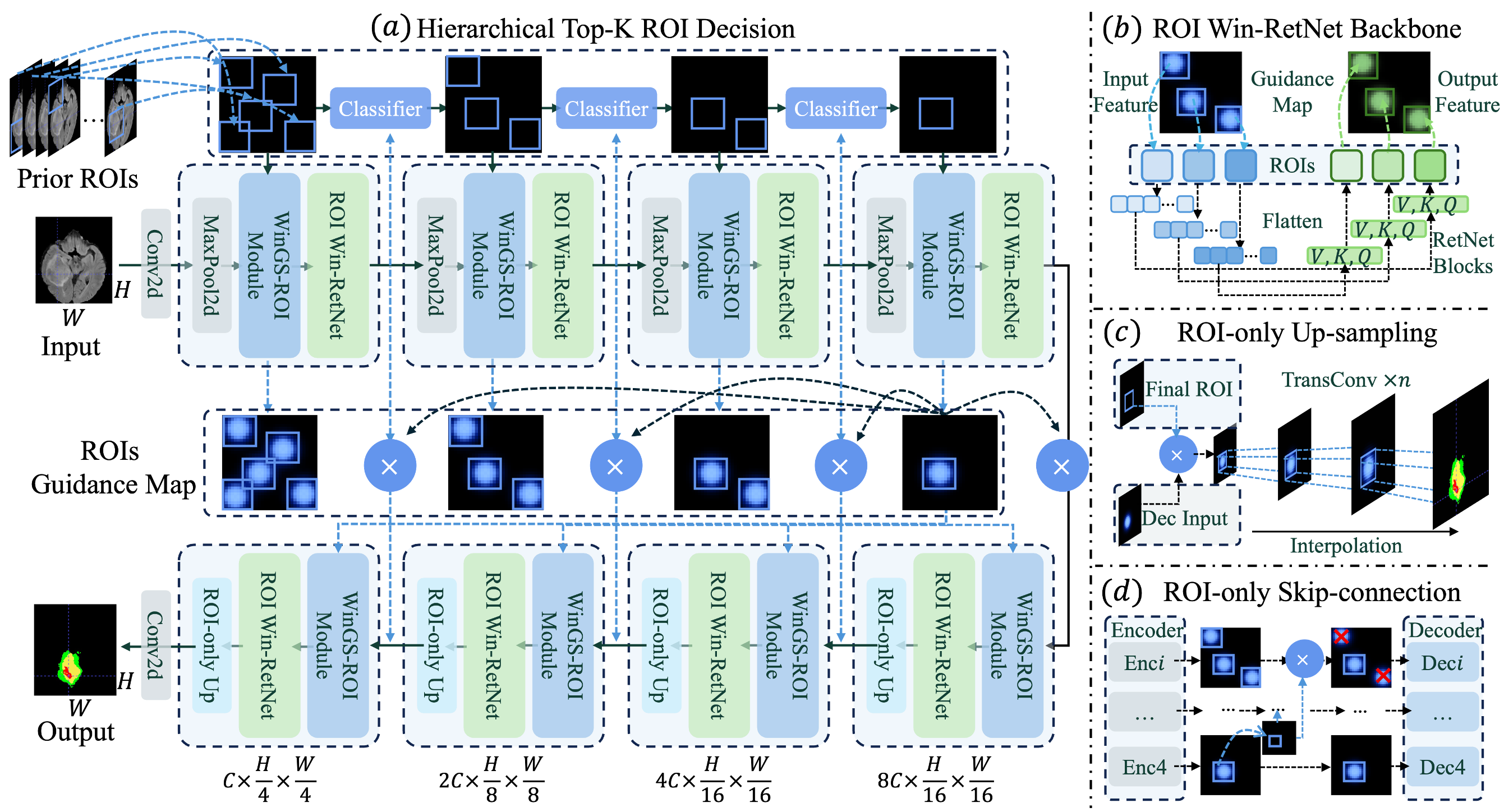}
%%    \caption{Overall architecture of \textbf{PGR-Net} (Prior-Guided Region Network). Including (a) Hierarchical Top-$K$ ROI decision, (b) ROI Win-RetNet, (c) ROI-only upsampling, and (d) Skip connections. The WinGS-ROI is shown in Fig.~\ref{fig:wings-roi}.}
    \caption{Overall architecture of \textbf{PGR-Net} (Prior-Guided ROI Reasoning Network), which consists of (a) Hierarchical Top-$K$ ROI Decision, (b) ROI Win-RetNet Backbone, (c) ROI-Only Up-sampling, and (d) ROI-Aware Skip-connections. The WinGS-ROI module is shown in Fig.~\ref{fig:wings-roi}}
    \label{fig:prnet}
   
\end{figure*}

\begin{figure*}[htbp]
    \centering
    \includegraphics[width=0.75\linewidth]{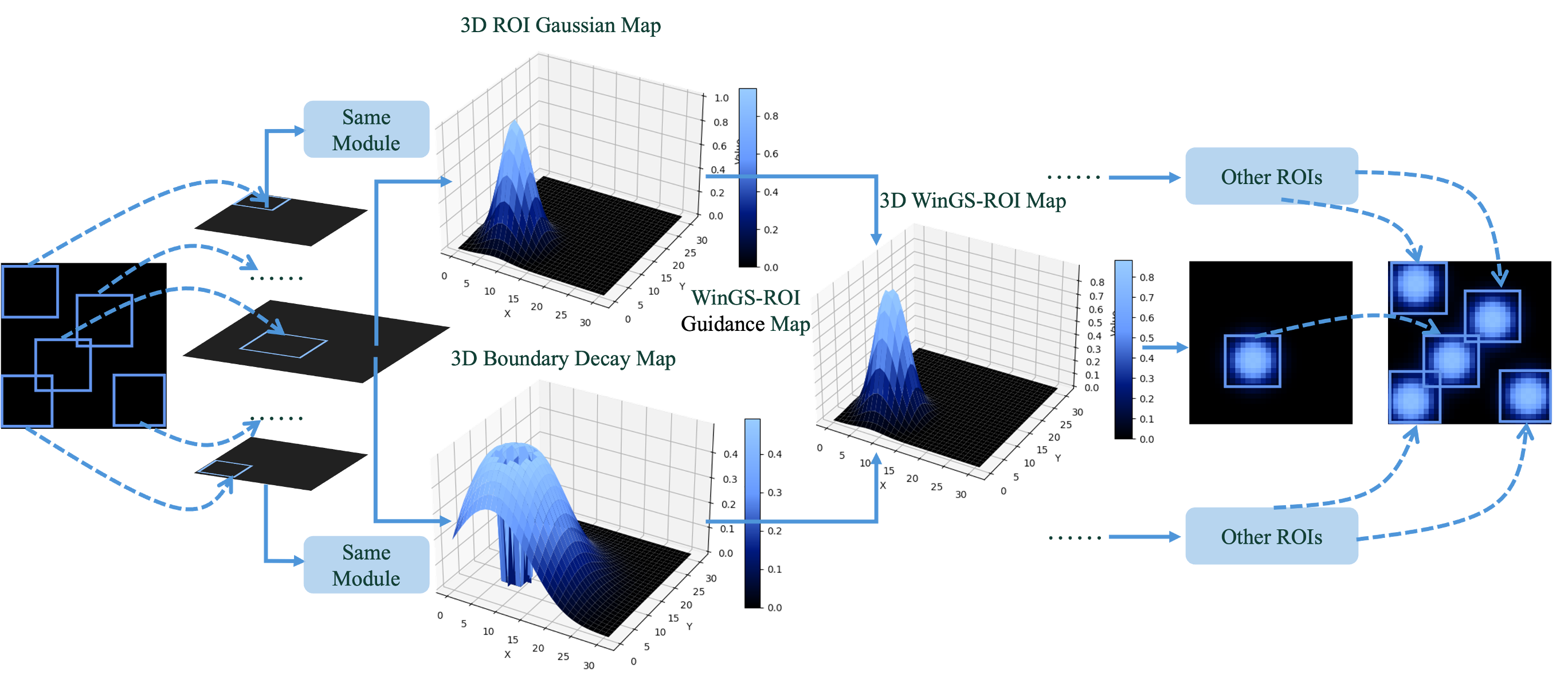}
    \caption{Illustration of the computation process of the WinGS-ROI mechanism. For each ROI region, a Gaussian template is applied within the ROI and a spatial decay is applied along the outer boundary, and the fused guidance map is then applied to the input image.}
    \label{fig:wings-roi}
\end{figure*}

Based on statistical priors, we design PGR-Net (Prior-Guided ROI Reasoning Network, see Fig.\ref{fig:prnet}) to achieve hierarchical ROI-aware segmentation from global to local scales. The network employs a windowed RetNet as the visual backbone to model regions within ROIs (see Fig.\ref{fig:prnet}(b)) and introduces a hierarchical Top-$K$ ROI (HTK, see Fig.\ref{fig:prnet}(a)) mechanism to progressively localize targets. Windowed Gaussian–Spatial Decay ROI (WinGS-ROI, see Fig.\ref{fig:wings-roi}) templates are embedded in feature modeling at each layer as learnable explicit spatial priors.

All ROI guidance is generated via a unified WinGS-ROI mechanism, ensuring consistent lesion focus and suppression of background regions. In shallow layers, where the ROI is not yet determined, the module operates in a multi-ROI candidate mode; once the ROI is locked, it switches to a single-ROI mode. During decoding, an ROI-only strategy restricts skip-connections and up-sampling to ROI regions, reducing background interference (Fig.~\ref{fig:prnet}(c)-(d)). ROI-Only up-sampling operates only inside the ROI with zero padding elsewhere, while ROI-Aware skip-connection propagates encoder features only within the ROI. The ROI used in decoding is the final selected region rather than intermediate candidates, ensuring consistent guidance.

\subsubsection{Hierarchical Top-$K$ (HTK) ROI Decision}

We propose the Hierarchical Top-$K$ (HTK) mechanism for dynamic and hierarchical ROI selection during inference. HTK recursively filters candidate ROIs across encoding layers, progressively searching for the final accurate ROI to enable subsequent ROI-only segmentation.

Let $\{(r_i, c_i)\}_{i=1}^N$ denote ROI prior templates. At layer $l$, each ROI generates a feature window:
\begin{equation}
\text{ROI}_i^{(l)} = \text{ROI}(r_i, c_i, F^{(l)}),
\end{equation}
where $F^{(l)}$ is the encoded feature map.

\paragraph{Top-$K$ candidate selection:}  
At the coarsest layer ($l=L$), ROI scores are computed by a lightweight MLP $f_\theta^{(L)}$:
\begin{equation}
s_i^{(L)} = f_\theta^{(L)}(F^{(L)},\, \text{ROI}_i^{(L)}), \quad i=1,\ldots,N,
\end{equation}
and the top $K^{(L)}$ candidates are selected:
\begin{equation}
\mathcal{T}^{(L)} = \text{TopK}(\{s_i^{(L)}\},\, K^{(L)}).
\end{equation}

At lower layers ($l=L-1,\dots,1$), scores are recomputed only for previously selected candidates:
\begin{equation}
s_i^{(l)} = f_\theta^{(l)}(F^{(l)},\, \text{ROI}_i^{(l)}), \quad i \in \mathcal{T}^{(l+1)},
\end{equation}
producing local decision vectors:
\begin{equation}
\mathbf{s}^{(l)} = [s_i^{(l)}]_{i \in \mathcal{T}^{(l+1)}}.
\end{equation}
\paragraph{Full-layer confidence matrix:}  
To ensure cross-layer comparability, define a global ROI index $1\!:\!N$. Each layer's scores are expanded:
\begin{equation}
\hat{s}_i^{(l)} =
\begin{cases}
\text{softmax}(\mathbf{s}^{(l)})_i, & i \in \mathcal{T}^{(l)},\\
0, & \text{otherwise},
\end{cases}
\end{equation}
and the full-layer confidence matrix is:
\begin{equation}
\mathbf{S} = \sum_{l=1}^{L} \alpha_l \cdot \hat{\mathbf{s}}^{(l)},
\end{equation}
where $\alpha_l$ is the layer weight. The final ROI decision is then:
\begin{equation}
R^* = \arg\max_i \mathbf{S}_i.
\end{equation}
\paragraph{Decision stability:}  
To prevent unreliable selections, confidence gap and entropy criteria are introduced:
\begin{equation}
\begin{aligned}
\Delta_{\text{gap}} &= s_{\text{top1}} - s_{\text{top2}}, \\
H &= -\sum_i p_i \log p_i, \quad
p_i = \frac{\exp(\mathbf{S}_i)}{\sum_j \exp(\mathbf{S}_j)}.
\end{aligned}
\end{equation}
If $\Delta_{\text{gap}} < \tau_1$ or $H > \tau_2$, the decision is considered uncertain, and the model falls back to full-image mode to avoid error propagation.

HTK operates synchronously with all encoding layers and is trained end-to-end together with PGR-Net through the segmentation loss, enabling dynamic, hierarchical, and reliable ROI selection throughout the network.

\subsubsection{ROI Win-RetNet Backbone}

The proposed ROI Win-RetNet (see Fig.~\ref{fig:prnet}(b)) serves as the core visual backbone of PGR-Net. It leverages the guidance map $M_{\text{roi}} \in \mathbb{R}^{H \times W}$ from the WinGS-ROI module to determine dynamic ROI windows and attention centers, enabling adaptive enhancement of lesion regions.

Given an input feature map $F^{(l)} \in \mathbb{R}^{C_l \times H_l \times W_l}$ and ROI prior templates $\{(r_i, c_i)\}_{i=1}^N$, the Top-$K$ candidate ROIs are selected according to the HTK full-layer confidence matrix $S$:
\begin{equation}
\mathcal{R}^{(l)} = \{ R_k^{(l)} = (c_k^{(l)}, r_k^{(l)}, \rho_k^{(l)}) \mid k = 1, \dots, K_l \},
\end{equation}
where $c_k^{(l)}$ and $r_k^{(l)}$ denote the normalized center and scale ratio of the ROI, and $\rho_k^{(l)}$ is the confidence score from HTK.

For each ROI, the corresponding window is extracted from $F^{(l)}$ based on $(c_k^{(l)}, r_k^{(l)})$ and flattened into a sequence:
\begin{equation}
X_k^{(l)} = \{ x_t^{(l)} \}_{t=1}^{N_k}.
\end{equation}
A RetNet block \cite{retnet} is applied to model sequential dependencies within each ROI window:
\begin{equation}
\{ h_t^{(l)} \}_{t=1}^{N_k} = \text{RetNet}(X_k^{(l)}),
\end{equation}

where $h_t^{(l)}$ represents the hidden states after RetNet propagation, capturing long-range dependencies efficiently.

Finally, outputs from all ROI windows are fused using confidence-weighted aggregation:
\begin{equation}
Y^{(l)} = \sum_{k=1}^{K_l} \omega_k^{(l)} \cdot \text{Fusion}(h_k^{(l)}), \quad
\omega_k^{(l)} = \frac{\exp(\gamma \rho_k^{(l)})}{\sum_j \exp(\gamma \rho_j^{(l)})},
\end{equation}
where $\gamma$ controls the sharpness of the weighting, ensuring globally consistent modeling of regional dependencies while emphasizing high-confidence ROIs.

This design aligns with the ROI priors, HTK selection, and WinGS-ROI guidance, enabling the Win-RetNet backbone to process each candidate ROI window via RetNet and aggregate local features into a global feature map for decoding.

\subsection{WinGS-ROI Mechanism}

\begin{figure}[htbp]
    \centering
    \includegraphics[width=1\linewidth]{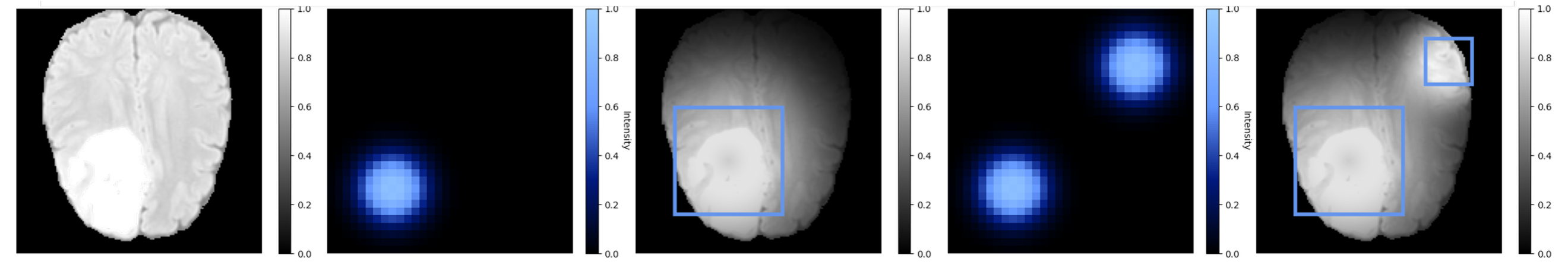}
    \caption{Example results after WinGS-ROI guidance. From left to right are the input image example, the WinGS-ROI guidance map of a single ROI, the WinGS-ROI guided image of a single ROI, the WinGS-ROI guidance map of dual ROIs, and the WinGS-ROI guided image of dual ROIs.}
    \label{fig:wings-sample}
\end{figure}

To enhance spatial sensitivity and boundary response, PGR-Net applies the WinGS-ROI (Windowed Gaussian–Spatial Decay ROI) mechanism at each layer. It builds a Gaussian template for each ROI with center enhancement and smooth boundary decay (Fig.\ref{fig:wings-roi}), focusing on lesion interiors while gradually suppressing edges, reducing background interference and preserving structural continuity. An example is shown in Fig.\ref{fig:wings-sample}.

\subsubsection{Gaussian Template Modeling}
Each ROI is now represented as a circular Gaussian template centered at $c_i$ with standard deviation $\sigma_i$, modulated by the HTK confidence $\rho_i^{(l)}$.  
Given candidate ROIs $\{(r_i, c_i, \sigma_i)\}_{i=1}^{K_l}$ at layer $l$, the template is:
\begin{equation}
G_i^{(l)}(u,v) = \rho_i^{(l)} \exp\!\Big(-\frac{(u - x_i)^2 + (v - y_i)^2}{2\sigma_i^2}\Big),
\end{equation}
where $(x_i, y_i)$ are the coordinates of $c_i$. This circular Gaussian ensures consistency with the ROI prior and provides a smooth, continuous saliency map highlighting the lesion center.

\subsubsection{Boundary-Aware Spatial Decay}
To suppress responses outside the ROI, define the radial distance to the ROI center:
\begin{equation}
d_i(u,v) = \sqrt{(u - x_i)^2 + (v - y_i)^2}.
\end{equation}
The decayed template is:
\begin{equation}
\begin{aligned}
\tilde{G}_i^{(l)}(u,v) =\\
\begin{cases}
G_i^{(l)}(u,v), & d_i(u,v) \le R_i,\\[1mm]
G_i^{(l)}(u,v) \exp\!\Big(-\frac{(d_i(u,v)-R_i)^2}{2\tau^2}\Big), & d_i(u,v) > R_i,
\end{cases}
\end{aligned}
\end{equation}
where $R_i$ is the radius corresponding to the ROI scale $r_i$, ensuring circular shape consistency, and $\tau$ controls the decay rate.  

The layer-wise guidance map aggregates all ROIs with confidence weighting:
\begin{equation}
M^{(l)}(u,v) = \frac{\sum_{i=1}^{K_l} \tilde{G}_i^{(l)}(u,v)}{\sum_{i=1}^{K_l} 1 + \epsilon},
\end{equation}
providing an adaptive, spatially weighted map for feature modulation.

\subsubsection{Multiplicative Modulation}
The original feature map $F^{(l)}$ is modulated by the guidance map:
\begin{equation}
\tilde{F}^{(l)} = (1 + \lambda M^{(l)}) \odot F^{(l)},
\end{equation}
where $\lambda$ balances enhancement and background preservation, and $\odot$ denotes element-wise multiplication.  

When the ROI is confidently locked ($\Delta_{\text{gap}} > \tau_{\text{lock}}$), a hard circular mask is applied:
\begin{equation}
\begin{aligned}
\tilde{F}^{(l)}(u,v) =\\
\begin{cases}
(1 + \lambda M^{(l)}(u,v)) F^{(l)}(u,v), & d_i(u,v) \le R_i,\\[1mm]
0, & \text{otherwise}.
\end{cases}
\end{aligned}
\end{equation}
The modulated ROI features are fused with the windowed RetNet output:
\begin{equation}
F_{\text{out}}^{(l)} = \hat{F}^{(l)} + \tilde{F}^{(l)},
\end{equation}

where $\hat{F}^{(l)}$ is the Win-RetNet output and $\tilde{F}^{(l)}$ is the guidance-modulated feature.  
When the ROI is confidently locked, only $\hat{F}^{(l)}$ is used:
\begin{equation}
F_{\text{out}}^{(l)} = \hat{F}^{(l)}.
\end{equation}
\section{Experiments}

\subsection{Datasets}
BraTS-2019 and BraTS-2023\cite{data1, data2, data3} are publicly available brain tumor datasets from MICCAI 2019 and 2023. MSD-Task01\cite{msd1, msd2} is part of the Medical Segmentation Decathlon. Each BraTS MRI volume has a volume resolution of \(155 \times 240 \times 240\) with four modalities. Voxel labels include: 0 (background), 1 (necrotic/non-enhancing), 2 (edema), and 4 (enhancing). Segmentation targets follow the BraTS convention: ET (4), TC (1+4), and WT (1+2+4). MSD-Task01 uses label 3 for ET.

\begin{table}[ht]
\centering
\caption{Modalities and the number of 2D slices for each dataset.}
\renewcommand\arraystretch{1.3}
    \footnotesize
\label{tab:data-summary}
% \resizebox{1\linewidth}{!}{
\begin{tabular}{lcc}
\hline
Dataset & Slices & Modalities \\
\hline
BraTS-2019 & 51925 & T1/T1ce/T2/FLAIR  \\
BraTS-2023 & 193905 & T1/T1ce/T2/FLAIR  \\
MSD-Task01 & 75020 & T1/T1ce/T2/FLAIR  \\
\hline
\end{tabular}
% }
\end{table}

Under computational constraints, all 3D volumes were sliced along the height axis into 2D images, and all models were implemented in 2D form. The resulting numbers of 2D slices for the three datasets are summarized in Table~\ref{tab:data-summary}. All datasets were split into training and testing sets at the case level with an 8:2 ratio. To reduce background redundancy, black background regions were cropped to \(160 \times 160\). Z-score normalization\cite{zscore} was further applied to the foreground to mitigate intensity variations across modalities.

% BraTS-2019 and BraTS-2023~\cite{data1, data2, data3} are publicly available brain tumor datasets released by MICCAI 2019 and 2023, comprising 335 and 1251 training cases, respectively. Each MRI volume has a spatial resolution of \(155 \times 240 \times 240\) and includes four modalities: T1, T1c, T2, and FLAIR. The voxel-wise labels are defined as 0 for non-tumor, 1 for necrotic/non-enhancing tumor, 2 for edema, and 4 for enhancing tumor. The segmentation targets are defined as Enhancing Tumor (ET, label 4), Tumor Core (TC, labels 1+4), and Whole Tumor (WT, labels 1+2+4).  

% MSD-Task01 (Brain Tumor)~\cite{msd1, msd2}, a subset of the Medical Segmentation Decathlon (MSD), consists of multi-modal MRI scans from 484 patients, each containing four modalities (T1, T1Gd, T2, and FLAIR) with voxel-level annotations. Unlike BraTS, the Enhancing Tumor (ET) in MSD-Task01 is labeled as 3.

% We split the datasets into training and testing sets with a ratio of 8:2. To alleviate the imbalance between background and target regions, black background areas in the MRI volumes were cropped, resulting in a volume size of 155 \(\times 160 \times 160\) per image. Due to large contrast variations among MRI modalities, Z-score normalization was applied to the foreground regions of each image \cite{zscore}. Considering computational limitations, all processed 3D images were sliced along the height axis into 2D images. All subsequent algorithms were implemented in their 2D versions accordingly.

\begin{table*}[htbp]
\centering

\caption{Ablation study results of PGR-Net on the BraTS-2019/2023 dataset. 
RWR: ROI Win-RetNet; HTK: Hierarchical Top-$K$ ROI Decision; 
WinGS-ROI includes WR: WinGS-ROI in Win-RetNet; SC: WinGS-ROI in Skip-connection; UP: WinGS-ROI in Up-sampling. 
Dice (\%) $\uparrow$ and Hausdorff95 (HD95) $\downarrow$ are reported for WT, TC, and ET regions.}
\renewcommand\arraystretch{1.3}
\footnotesize
\resizebox{\linewidth}{!}{
\begin{tabular}{c|cc|ccc|ccc|ccc}
% \hline
\toprule
Model & \multicolumn{2}{c|}{Backbone} & \multicolumn{3}{c|}{WinGS-ROI} & \multicolumn{3}{c|}{Dice\_score (\%)} & \multicolumn{3}{c}{Hausdorff95} \\
\cline{2-12}
 & RWR & HTK & WR & SC & UP & WT↑ & TC↑ & ET↑ & WT↓ & TC↓ & ET↓ \\
\hline
A (Baseline) & - & - & - & - & - & 87.82 / 91.06 & 88.91 / 92.97 & 91.05 / 93.13 & 1.3264 / 1.1868 & 0.8409 / 0.7085 & 0.6645 / 0.6622 \\
B & + & - & - & - & - & 87.85 / 91.10 & 88.89 / 93.02 & 91.15 / 93.08 & 1.3205 / 1.1840 & 0.8351 / 0.7055 & 0.6650 / 0.6605 \\
C & + & + & - & - & - & 88.55 / 91.66 & 89.64 / 93.42 & 91.99 / 93.35 & 1.2911 / 1.1551 & 0.8275 / 0.6803 & 0.6439 / 0.6419 \\
D & + & + & + & - & - & 88.63 / 91.76 & 90.33 / 93.75 & 92.72 / 93.57 & 1.2887 / 1.1457 & 0.8125 / 0.6754 & 0.6406 / 0.6327 \\
E & + & + & + & + & - & 88.85 / 91.80 & 90.32 / 93.79 & 92.88 / 93.74 & 1.2692 / 1.1380 & 0.8129 / 0.6703 & 0.6400 / 0.6153 \\
F (Full Model) & + & + & + & + & + & \textbf{89.02 / 91.82} & \textbf{90.69 / 94.07} & \textbf{93.61 / 93.88} & \textbf{1.2633 / 1.1334} & \textbf{0.7988 / 0.6647} & \textbf{0.6371 / 0.6011} \\
\hline
\end{tabular}
}

\label{tab:ablation}
\end{table*}

\subsection{Metrics and Implementation Details}
Our network is implemented using the PyTorch framework on Ubuntu 22.04, with all experiments conducted on an NVIDIA RTX 2080Ti GPU. Evaluations related to computational cost and inference time are performed on the same device. Each experiment is independently executed three times to ensure statistical reliability, and the final results are reported as the mean of these runs. 

The loss function is a weighted combination of Dice loss\cite{diceloss} and BCE loss, with a weighting ratio of 2:8. All algorithms are trained for 300 epochs with an early stopping strategy of 50 epochs, using the Adam optimizer with an initial dynamic learning rate of 1e-3, ensuring fairness across all comparisons.

The performance of all algorithms is evaluated using the Dice Score and Hausdorff95 Distance. Dice Score quantifies the overlap between predicted and ground truth segmentations, widely used in medical image segmentation to evaluate segmentation consistency:
\begin{equation}
\text{Dice} = \frac{2TP}{FP + 2TP + FN}
\end{equation}
where \(TP\), \(FP\), and \(FN\) denote true positives, false positives, and false negatives, respectively.

HD95 measures the boundary discrepancy between two segmentation sets by computing the 95th percentile of the bidirectional surface distances:
\begin{equation}
\begin{aligned}
\text{Haus}(A,B) = \\\max \left( \max_{S_A \in S(A)} d(S_A, S(B)), \max_{S_B \in S(B)} d(S_B, S(A)) \right)
\end{aligned}
\end{equation}
where \(A\) and \(B\) are the two segmentation sets, \(d\) is the distance from an element to the closest point, and \(S(A)\) and \(S(B)\) represent the sets of elements in \(A\) and \(B\), respectively.

\subsection{Ablation Study}

To verify the effectiveness of each module, we conducted systematic ablation experiments on the BraTS 2019 and BraTS 2023 datasets. Each model was independently trained and evaluated three times, and the metric deviations across runs were within 0.06(Dice) and 0.002(HD95), indicating strong experimental consistency. Starting from the baseline model, we progressively introduced the ROI Win-RetNet (RWR), Hierarchical Top-$K$ ROI Decision (HTK), and the hierarchically embedded WinGS-ROI modules. The results are summarized in Table \ref{tab:ablation}, bold numbers in the table indicate the best performance.

The baseline model A contains only a basic encoder–decoder structure and exhibits limited performance. When the RWR module is added (model B), Dice scores show slight improvements across most regions. However, due to the absence of hierarchical selection and explicit spatial guidance, the performance gains remain modest. Model C further incorporates the HTK module, which dynamically selects the most representative ROI regions across multiple feature layers, enabling more precise spatial localization. This demonstrates that hierarchical selection effectively enhances ROI localization accuracy and improves boundary consistency.

In models D–F, the WinGS-ROI module is gradually introduced: when embedded within Win-RetNet (model D), it produces smoother boundary predictions; further integrating it into the Skip-connection (model E) and Up-sampling (model F) stages continually improves segmentation accuracy. The final full model (F) achieves the best performance (WT: 91.82, TC: 94.07, ET: 93.88) with the lowest HD95 values.

Notably, ROI guidance in PGR-Net acts as a soft constraint rather than a hard restriction. To ensure robustness, a fallback mechanism reverts to full-image processing when necessary. The fallback is triggered in 6.97\%, 3.52\%, and 5.33\% of cases on BraTS 2019, BraTS 2023, and MSD Task01, respectively, mainly for samples with abnormal morphology or distribution shifts.

Overall, these results indicate that RWR and HTK effectively construct stable ROI representations and hierarchical decisions, while the multi-stage WinGS-ROI guidance further strengthens spatial sensitivity and boundary delineation, leading to significant improvements in overall brain tumor segmentation performance.

\subsection{Comparison with the state-of-the-art methods}

\begin{table*}[ht]
    \centering
    \caption{
		Comparison with The SOTA methods on BraTS-2019 / BraTS-2023 / MSD Task01 Datasets.
	}
    \resizebox{\linewidth}{!}{
    \renewcommand\arraystretch{1.3}
    \footnotesize
    \begin{tabular}{c|c|ccc|ccc}
    \toprule
    % \hline
        \multirow{2}{*}{Model} & \multirow{2}{*}{Year} & \multicolumn{3}{c}{Dice\_score (\%)} & \multicolumn{3}{c}{Hausdorff95} \\ \cline{3-8} 
		&& WT↑ & TC↑ & ET↑ & WT↓ & TC↓ & ET↓ \\ \hline
        UNet & 2015 & 87.36 / 90.71 / 88.15 & 88.59 / 93.05 / 88.56 & 90.69 / 93.36 / 90.33 & 1.3582 / 1.1863 / 1.3800 & 0.9076 / 0.7329 / 0.9356 & 0.6897 / 0.6730 / 0.7861 \\ \hline
        Cascaded UNet & 2019& 87.81 / 90.32 / 89.06 & 89.40 / 92.85 / 88.29 & 90.92 / 92.47 / 90.24 & 1.3349 / 1.2091 / 1.3510 & 0.9002 / 0.7488 / 0.9268 & 0.6719 / 0.7591 / 0.7648 \\ \hline
        TransUNet & 2021 &  84.50 / 90.71 / 87.26 & 86.72 / 92.52 / 88.58 & 88.39 / 92.92 / 89.52 & 1.3911 / 1.1810 / 1.3936 & 0.9300 / 0.7276 / 0.9349 & 0.7396 / 0.6869 / 0.8073 \\ \hline
        nnUNet & 2021& 87.81 / 90.34 / 89.33 & 90.23 / 92.74 / 88.79 & 90.96 / 92.37 / 90.41 & 1.2970 / 1.2100 / 1.3420& 0.8311 / 0.7358 / 0.9250 & 0.6628 / 0.6722 / 0.7849 \\ \hline
        UNETR & 2022 & 85.29 / 88.35 / 85.92 & 87.16 / 89.16 / 85.82 & 89.54 / 91.43 / 88.52 & 1.3831 / 1.2427 / 1.4480 & 0.9504 / 0.8926 / 1.0447 & 0.7042 / 0.7211 / 0.8310 \\ \hline
        Swin UNETR & 2022 & 88.16 / 91.11 / 89.20 & 88.85 / 93.20 / 88.51 & 90.86 / 93.42 / 89.97 & 1.3077 / 1.1629 / 1.3571 & 0.9119 / 0.7088 / 0.9372 & 0.6814 / 0.6631 / 0.7989 \\ \hline
        SLf-UNet & 2024 & 87.55 / 90.81 / 88.20 & 88.21 / 93.18 / 88.61 & 90.38 / 93.30 / 90.46 & 1.3273 / 1.1748 / 1.3954 & 0.9032 / 0.7100 / 0.9533 & 0.6871 / 0.6709 / 0.7649\\ \hline
        MedSAM & 2024 & 85.39 / 88.55 / 84.20 & 87.90 / 91.55 / 86.11 & 88.20 / 90.30 / 86.72 & 1.4409 / 1.3155 / 1.5697 & 0.9224 / 0.8003 / 1.0025 & 0.7667 / 0.8153 / 0.9206 \\ \hline
        Mamba-UNet & 2024& 88.21 / 91.03 / 88.75 & 90.11 / 93.32 / 88.38 & 90.86 / 93.31 / 89.90 & 1.3061 / 1.1734 / 1.3459 & 0.8235 / 0.7008 / 0.9280 & 0.6750 / 0.6764 / 0.7879 \\ \hline
        UKAN & 2024 & 87.39 / 90.64 / 87.81 & 89.50 / 93.04 / 88.10 & 91.20 / 93.14 / 90.35 & 1.2989 / 1.1862 / 1.3410 & 0.8415 / 0.7234 / 0.9355 & 0.6585 / 0.6824 / 0.7880\\ \hline
        VM-UNet & 2024 & 87.74 / 90.52 / 89.05 & 90.39 / 93.40 / 88.43 & 91.06 / 93.50 / 89.76 & 1.3122 / 1.1806 / 1.3393 & 0.8258 / 0.7079 / 0.9250 & 0.6744 / 0.6781 / 0.7912 \\ \hline
        M-Net & 2025 & 88.38 / 91.33 / 89.04 & 90.52 / 93.55 / 88.55 & 91.43 / 93.42 / 90.26 & 1.2869 / 1.1534 / 1.3359 & 0.8154 / 0.7069 / 0.9256 & 0.6571 / 0.6600 / 0.7701 \\ \hline
        H-VMUNet & 2025 & 87.95 / 90.77 / 89.12 & 90.03 / 93.05 / 88.59 & 90.50 / 92.88 / 90.14 & 1.3104 / 1.1833 / 1.3341 & 0.8411 / 0.7229 / 0.9244 & 0.6821 / 0.6931 / 0.7789 \\ \hline
        Mamba Sea & 2025 & 88.19 / 91.12 / 89.15 & 90.19 / 93.41 / 88.40 & 91.44 / 93.47 / 90.28 & 1.3081 / 1.1837 / 1.3398 & 0.8152 / 0.6977 / 0.9352 & 0.6569 / 0.6728 / 0.7735\\ \hline

        PGR-Net(ours) & -- & \textbf{89.02 / 91.82 / 89.67} & \textbf{90.69 / 94.07 / 89.27} & \textbf{93.61 / 93.88 / 90.63} & \textbf{1.2633 / 1.1334 / 1.3144} & \textbf{0.7988 / 0.6647 / 0.9089} & \textbf{0.6371 / 0.6011 / 0.7571} \\
        
        \hline
    \end{tabular}
    }
    
	\label{table:com}
\end{table*}

\begin{figure*}[ht]
	\centering
	\includegraphics[width=0.95\textwidth]{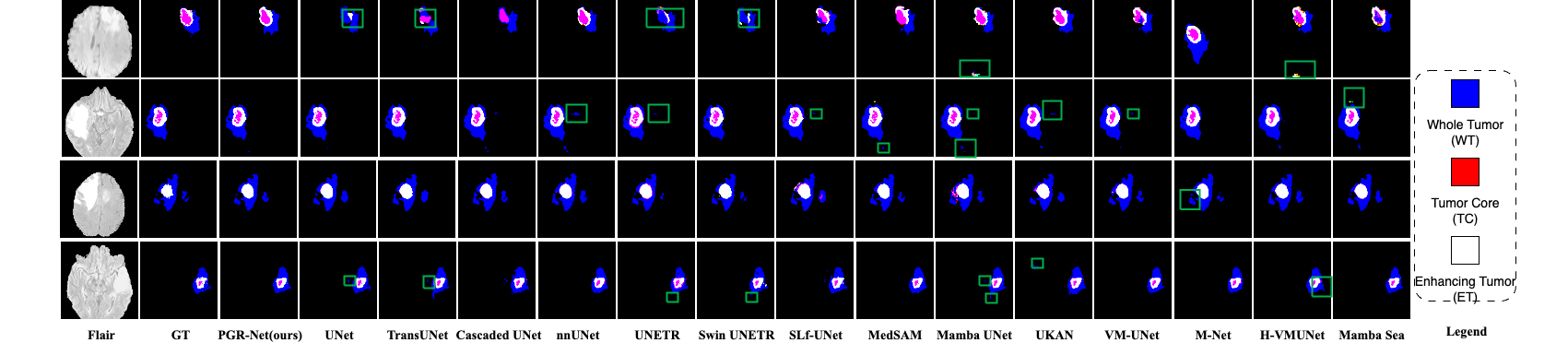} 
	\caption{Examples of segmentation results from multiple methods. From left to right: Flair modality input image, Ground Truth (GT), and segmentation results from various comparison algorithms. Green boxes highlight regions with obvious segmentation errors.}
	\label{fig:com}
\end{figure*}

To thoroughly evaluate the effectiveness and superiority of the proposed method, we conducted comparative experiments on three brain tumor segmentation datasets: BraTS 2019, BraTS 2023, and MSD Task01.  
We compared with convolution-based networks (UNet\cite{Ron1}, Cascaded UNet\cite{casunet}), Transformer-based architectures (TransUNet\cite{transunet}, UNETR\cite{unetr}, Swin UNETR\cite{swin}), and recent state-space model (SSM) methods (Mamba-UNet\cite{mambaunet}, VM-UNet\cite{vmamba}, M-Net\cite{mnet}, H-VMUNet\cite{hvmunet}), including their 2D variants whenever official implementations were available. All models were independently trained and tested three times under the same training configurations, and the results are averaged. Bold values in the tables indicate the best performance.

\begin{table}[ht]
\centering
\caption{Comparison of model parameters and computational complexity on the BraTS 2023 test set, computed using a batch size of 16.}
\resizebox{1\linewidth}{!}{
    \footnotesize
\begin{tabular}{c|c|ccc}
\toprule
Model & Year & Params(M) & FLOPs(G) & Inf Time(min) \\
\hline
UNet & 2015 & 39.40 & 321.19 & 12:32 \\ \hline
Cascaded UNet & 2019 & 85.59 & 568.25 & 25:24 \\ \hline
TransUNet & 2021 & 105.21 & 237.83 & 11:02 \\ \hline
nnUNet & 2021 & -- & -- & 86:52 \\ \hline
UNETR & 2022 & 87.67 & 150.71 & 18:31 \\ \hline
Swin UNETR & 2022 & 25.11 & 106.80 & 21:33 \\ \hline
SLf-UNet & 2024 & 36.08 & 534.73 & 30:19 \\ \hline
MedSAM & 2024 & 240.32 & 166.55 & 30:19 \\ \hline
Mamba-UNet & 2024 & 35.86 & 72.44 & 14:12 \\ \hline
UKAN & 2024 & 25.36 & 62.21 & 19:43 \\ \hline
VM-UNet & 2024 & 44.28 & 61.42 & 13:52 \\ \hline
M-Net & 2025 & 81.59 & 91.29 & 15:33 \\ \hline
H-VMUNet & 2025 & 51.92 & 75.98 & 16:04 \\ \hline
Mamba Sea & 2025 & 27.43 & 66.91 & 16:49 \\ \hline

PGR-Net(ours) & -- & \textbf{8.64} & \textbf{39.05} & \textbf{9:41} \\
\bottomrule
\end{tabular}
}

\label{tab:computation}
\end{table}

As shown in Table~\ref{table:com}, PGR-Net consistently achieves superior segmentation accuracy across all three datasets, demonstrating particularly remarkable improvements in the Whole Tumor (WT) region.
On the BraTS 2023 dataset, PGR-Net attains a WT Dice score of 91.82\%, outperforming state-of-the-art methods such as Swin UNETR, VM-UNet, and Mamba Sea by approximately 0.8–1.4\% on average, while also achieving significantly lower HD95 values.

In addition, as shown in the parameter and computational complexity comparison (Table~\ref{tab:computation}), PGR-Net contains only 8.64M parameters with 39.05G FLOPs, and requires just 9 minutes and 41 seconds for inference—substantially lower than other mainstream approaches.

The notable improvements stem from the core design philosophy of PGR-Net: Allocating computational resources to actual lesion areas, PGR-Net forms an accurate global tumor perception at early network stages, thereby maintaining strong coherence and cross-layer consistency in subsequent feature fusion and boundary refinement.

As shown in Fig.~\ref{fig:com}, qualitative results on the BraTS 2023 test set further demonstrate the advantages of PGR-Net. The proposed method produces more accurate boundaries and better preserves fine tumor structures, particularly maintaining the completeness and continuity of the WT region. These results validate the effectiveness of the proposed ROI-based prior modeling and hierarchical guidance mechanism for brain tumor MRI segmentation.

%%As illustrated in Fig.~\ref{fig:com}, qualitative comparisons on the BraTS 2023 test set further demonstrate that PGR-Net produces more precise boundary delineation and better reconstruction of fine tumor structures—especially maintaining the completeness and continuity of the WT region, which verifies the effectiveness and advantage of the proposed ROI-based prior modeling and hierarchical guidance mechanism in brain tumor MRI segmentation.

\section{Conclusion}
This work targets the persistent difficulties in brain tumor MRI segmentation, where lesions are sparsely distributed and the background dominates most of the volume. PGR-Net addresses these issues by introducing explicit spatial priors into the feature learning process rather than relying solely on appearance-driven attention mechanisms. The framework employs a data-derived ROI prior to inform two complementary components: the WinGS-ROI module, which embeds center-weighted spatial cues into windowed features, and a HTK ROI Decision branch that selects informative regions across layers in a consistent manner. This design enables the network to reason about potential lesion areas throughout feature propagation, aided by the introduced windowed RetNet backbone. Experimental results on BraTS 2019, BraTS 2023, and MSD Task01 consistently demonstrate that the proposed prior-guided strategy yields measurable benefits, particularly for Whole Tumor segmentation, while requiring only minimal computational resources.
Extending PGR-Net beyond WT priors to TC, ET, and other lesion regions could further improve multi-class spatial modeling and segmentation reliability.

\section*{Acknowledgments}
This work was supported by the National Natural Science Foundation of China (62476178), Beijing Natural Science Foundation of China (4242034),  and the National Key Laboratory of Human-Machine Hybrid Augmented Intelligence, Xi'an Jiaotong University (No. HMHAI-202407).

%% where spatial ambiguity is most substantial
%%This work currently uses WT priors, which could be extended in the future to TC, ET, and other lesion regions.
% This work targets the persistent difficulties in brain tumor MRI segmentation, where lesions are sparsely distributed, highly variable in shape, and unevenly spread across scales. PGR-Net addresses these issues by introducing explicit spatial priors into the feature learning process rather than relying solely on appearance-driven attention mechanisms. The framework employs a data-derived ROI prior to inform two complementary components: the WinGS-ROI module, which embeds center-weighted spatial cues into windowed features, and a Hierarchical Top-K ROI Decision branch that selects informative regions across layers in a consistent manner. This design enables the network to reason about potential lesion areas throughout feature propagation, improving localization reliability under challenging cases. Furthermore, PGR-Net exhibits lightweight and efficient characteristics in terms of parameter count and computational complexity. Experimental results on BraTS 2019, BraTS 2023, and MSD Task01 provide consistent evidence that the proposed prior-guided strategy yields measurable benefits, particularly for Whole Tumor segmentation where spatial ambiguity is most substantial.

{
    \small
    \balance

    \bibliographystyle{ieeenat_fullname}
    \bibliography{main}

@article{Tan1,
  author = {Tan, A. C. and Ashley, D. M. and López, G. Y. and Malinzak, M. and Friedman, H. S. and Khasraw, M.},
  title = {Management of glioblastoma: State of the art and future directions},
  journal = {CA: a cancer journal for clinicians},
  volume = {70},
  number = {4},
  pages = {299-312},
  year = {2020}
}

@book{Liang1,
  author = {Liang, Z. P. and Lauterbur, P. C.},
  title = {Principles of magnetic resonance imaging},
  pages = {1-7},
  year = {2000},
  publisher = {SPIE Optical Engineering Press},
  address = {Bellingham}
}

@article{Fuku,
  author = {Fukushima, K.},
  title = {Neocognitron: A self-organizing neural network model for a mechanism of pattern recognition unaffected by shift in position},
  journal = {Biological cybernetics},
  volume = {36},
  number = {4},
  pages = {193-202},
  year = {1980}
}

@inproceedings{Ron1,
  author = {Ronneberger, O. and Fischer, P. and Brox, T.},
  title = {U-net: Convolutional networks for biomedical image segmentation},
  booktitle = {Medical image computing and computer-assisted intervention–MICCAI 2015: 18th international conference, Munich, Germany, October 5-9, 2015, proceedings, part III 18},
  pages = {234-241},
  publisher = {Springer International Publishing},
  year = {2015}
}

@inproceedings{transformer,
  author    = {Ashish Vaswani and Noam Shazeer and Niki Parmar and Jakob Uszkoreit and Llion Jones and Aidan N. Gomez and Lukasz Kaiser and Illia Polosukhin},
  title     = {Attention Is All You Need},
  booktitle = {Advances in Neural Information Processing Systems},
  volume    = {30},
  year      = {2017}
}

@inproceedings{mamba,
  author    = {Albert Gu and Tri Dao},
  title     = {Mamba: Linear-Time Sequence Modeling with Selective State Spaces},
  booktitle = {First Conference on Language Modeling (COLM)},
  year      = {2024}
}

@article{transunet,
  title={TransUNet: Rethinking the U-Net architecture design for medical image segmentation through the lens of transformers},
  author={Chen, Jieneng and Mei, Jieru and Li, Xianhang and Lu, Yongyi and Yu, Qihang and Wei, Qingyue and Luo, Xiangde and Xie, Yutong and Adeli, Ehsan and Wang, Yan and others},
  journal={Medical Image Analysis},
  pages={103280},
  year={2024},
  publisher={Elsevier}
}

@inproceedings{unetr,
  author = {Hatamizadeh, A. and Tang, Y. and Nath, V. and Yang, D. and Myronenko, A. and Landman, B. and Xu, D.},
  title = {Unetr: Transformers for 3d medical image segmentation},
  booktitle = {Proceedings of the IEEE/CVF winter conference on applications of computer vision},
  pages = {574-584},
  year = {2022}
}

@inproceedings{swin,
  author = {Hatamizadeh, A. and Nath, V. and Tang, Y. and Yang, D. and Roth, H. R. and Xu, D.},
  title = {Swin unetr: Swin transformers for semantic segmentation of brain tumors in mri images},
  booktitle = {International MICCAI brainlesion workshop},
  pages = {272-284},
  publisher = {Cham: Springer International Publishing},
  year = {2021}
}

@article{vmamba,
  author    = {Zhu, L. and Liao, B. and Zhang, Q. and Wang, X. and Liu, W. and Wang, X.},
  title     = {Vision mamba: Efficient visual representation learning with bidirectional state space model},
  journal   = {arXiv preprint},
  year      = {2024},
  eprint    = {2401.09417}
}

@article{mambaunet,
  author    = {Wang, Z. and Zheng, J. Q. and Zhang, Y. and Cui, G. and Li, L.},
  title     = {Mamba-unet: Unet-like pure visual mamba for medical image segmentation},
  journal   = {arXiv preprint},
  year      = {2024},
  eprint    = {2402.05079}
}

@inproceedings{diceloss,
  author    = {Xiaoya Li and Xiaofei Sun and Yuxian Meng and Junjie Liang and Fei Wu and Jiwei Li},
  title     = {Dice Loss for Data-Imbalanced NLP Tasks},
  booktitle = {Proceedings of the 58th Annual Meeting of the Association for Computational Linguistics (ACL)},
  pages     = {465--476},
  year      = {2020}
}

@article{data1,
  author    = {Menze, B. H. and Jakab, A. and Bauer, S. and Kalpathy-Cramer, J. and Farahani, K. and Kirby, J. and Van Leemput, K.},
  title     = {The multimodal brain tumor image segmentation benchmark (BRATS)},
  journal   = {IEEE transactions on medical imaging},
  volume    = {34},
  number    = {10},
  pages     = {1993--2024},
  year      = {2014}
}

@article{data2,
  author    = {Bakas, S. and Akbari, H. and Sotiras, A. and Bilello, M. and Rozycki, M. and Kirby, J. S. and Davatzikos, C.},
  title     = {Advancing the cancer genome atlas glioma MRI collections with expert segmentation labels and radiomic features},
  journal   = {Scientific data},
  volume    = {4},
  number    = {1},
  pages     = {1--13},
  year      = {2017}
}

@article{data3,
  author    = {Bakas, S. and Reyes, M. and Jakab, A. and Bauer, S. and Rempfler, M. and Crimi, A. and Jambawalikar, S. R.},
  title     = {Identifying the best machine learning algorithms for brain tumor segmentation, progression assessment, and overall survival prediction in the BRATS challenge},
  journal   = {arXiv preprint},
  year      = {2018},
  eprint    = {1811.02629}
}

@article{liu2024vmamba,
  author    = {Liu, Yujun and Tian, Yuxin and Zhao, Yujie and others},
  title     = {VMamba: Visual State Space Model},
  journal   = {Advances in Neural Information Processing Systems},
  volume    = {37},
  pages     = {103031--103063},
  year      = {2024}
}

@inproceedings{mnet,
  title={M-Net: MRI Brain Tumor Sequential Segmentation Network via Mesh-Cast},
  author={Lu, J. and Ding, H. and Zhang, S. and others},
  booktitle={Proceedings of the IEEE/CVF International Conference on Computer Vision},
  year={2025},
  pages={20116--20125}
}

@article{cli1,
  title={Incidence of gliomas by anatomic location},
  author={Larjavaara, S. and M{\"a}ntyl{\"a}, R. and Salminen, T. and others},
  journal={Neuro-Oncology},
  year={2007},
  volume={9},
  number={3},
  pages={319--325}
}

@article{cli2,
  title={Anatomical distribution and prognostic heterogeneity in glioma: unique clinical features of occipital glioblastoma},
  author={Zhao, C. and Liang, B. and Li, X. and others},
  journal={Journal of Neuro-Oncology},
  year={2025},
  pages={1--13}
}

@article{cli3,
  title={Anatomical features of primary brain tumors affect seizure risk and semiology},
  author={Akeret, K. and Serra, C. and Rafi, O. and others},
  journal={NeuroImage: Clinical},
  year={2019},
  volume={22},
  pages={101688}
}

@article{retnet,
  title={Retentive Network: A Successor to Transformer for Large Language Models},
  author={Sun, Y. and Dong, L. and Huang, S. and others},
  journal={arXiv preprint arXiv:2307.08621},
  year={2023}
}

@inproceedings{rmt,
  title={RMT: Retentive Networks Meet Vision Transformers},
  author={Fan, Q. and Huang, H. and Chen, M. and others},
  booktitle={Proceedings of the IEEE/CVF Conference on Computer Vision and Pattern Recognition},
  year={2024},
  pages={5641--5651}
}

@article{cas1,
  title={Cascaded 3D UNet architecture for segmenting the COVID-19 infection from lung CT volume},
  author={Aswathy, A. L. and SS, V. C.},
  journal={Scientific Reports},
  year={2022},
  volume={12},
  pages={3090}
}

@article{cas2,
  title={EG-Unet: Edge-Guided Cascaded Networks for Automated Frontal Brain Segmentation in MR Images},
  author={Zhang, X. and Liu, Y. and Guo, S. and others},
  journal={Computers in Biology and Medicine},
  year={2023},
  volume={158},
  pages={106891}
}

@inproceedings{cas3,
  title={Glioma segmentation with cascaded UNet},
  author={Lachinov, D. and Vasiliev, E. and Turlapov, V.},
  booktitle={International MICCAI Brainlesion Workshop},
  year={2018},
  pages={189--198},
  publisher={Springer International Publishing},
  address={Cham}
}

@article{attention1,
  title={Attention-UNet Architectures with Pretrained Backbones for Multi-Class Cardiac MR Image Segmentation},
  author={Das, N. and Das, S.},
  journal={Current Problems in Cardiology},
  year={2024},
  volume={49},
  number={1},
  pages={102129}
}

@inproceedings{attention2,
  title={Swin-UNet: UNet-like Pure Transformer for Medical Image Segmentation},
  author={Cao, H. and Wang, Y. and Chen, J. and others},
  booktitle={European Conference on Computer Vision},
  year={2022},
  pages={205--218},
  publisher={Springer Nature Switzerland},
  address={Cham}
}

@article{attention3,
  title={DS-TransUNet: Dual Swin Transformer U-Net for Medical Image Segmentation},
  author={Lin, A. and Chen, B. and Xu, J. and others},
  journal={IEEE Transactions on Instrumentation and Measurement},
  year={2022},
  volume={71},
  pages={1--15}
}

@article{roi1,
  title={A Radiomics-Incorporated Deep Ensemble Learning Model for Multi-Parametric MRI-Based Glioma Segmentation},
  author={Chen, Y. and Yang, Z. and Zhao, J. and others},
  journal={Physics in Medicine \& Biology},
  year={2023},
  volume={68},
  number={18},
  pages={185025}
}

@article{roi2,
  title={A Survey of MRI-Based Brain Tumor Segmentation Methods},
  author={Liu, J. and Li, M. and Wang, J. and others},
  journal={Tsinghua Science and Technology},
  year={2014},
  volume={19},
  number={6},
  pages={578--595}
}

@article{msd1,
  title={The Medical Segmentation Decathlon},
  author={Antonelli, Michela and Reinke, Annika and Bakas, Spyridon and others},
  journal={Nature Communications},
  year={2022},
  doi={10.1038/s41467-022-30695-9}
}

@misc{msd2,
  title={A Large Annotated Medical Image Dataset for the Development and Evaluation of Segmentation Algorithms},
  author={Simpson, Amber L. and Antonelli, Michela and Bakas, Spyridon and Bilello, Michel and Farahani, Keyvan and van Ginneken, Bram and Kopp-Schneider, Annette and Landman, Bennett A. and Litjens, Geert and Menze, Bjoern and Ronneberger, Olaf and Summers, Ronald M. and Bilic, Patrick and Christ, Patrick F. and Do, Richard K. G. and Gollub, Marc and Golia-Pernicka, Jennifer and Heckers, Stephan H. and Jarnagin, William R. and McHugo, Maureen K. and Napel, Sandy and Vorontsov, Eugene and Maier-Hein, Lena and Cardoso, M. Jorge},
  year={2019},
  eprint={1902.09063},
  archivePrefix={arXiv},
  primaryClass={cs.CV}
}

@inproceedings{casunet,
  title={Multi-Step Cascaded Networks for Brain Tumor Segmentation},
  author={Li, X. and Luo, G. and Wang, K.},
  booktitle={International MICCAI Brainlesion Workshop},
  year={2019},
  pages={163--173},
  publisher={Springer International Publishing},
  address={Cham}
}

@article{hvmunet,
  title={H-vmunet: High-Order Vision Mamba UNet for Medical Image Segmentation},
  author={Wu, R. and Liu, Y. and Liang, P. and others},
  journal={Neurocomputing},
  year={2025},
  volume={624},
  pages={129447}
}

@article{zscore,
  author    = {L. Al Shalabi and Z. Shaaban and B. Kasasbeh},
  title     = {Data Mining: A Preprocessing Engine},
  journal   = {Journal of Computer Science},
  year      = {2006},
  volume    = {2},
  number    = {9},
  pages     = {735--739},
  doi       = {},
  publisher = {},
  note      = {}
}
}

\end{document}